\documentclass[10pt,twocolumn,letterpaper]{article}

\usepackage{cvpr}              

\usepackage{amsmath}
\usepackage{amssymb}
\usepackage{amsfonts}
\usepackage{graphicx}
\usepackage{enumitem}
\usepackage{xcolor}

\definecolor{cvprblue}{rgb}{0.21,0.49,0.74}
\usepackage[pagebackref,breaklinks,colorlinks,allcolors=cvprblue]{hyperref}

\def\paperID{*****} 
\def\confName{CVPR}
\def\confYear{2026}

\title{Pixel Perfect: Relational Image Quality Assessment with Spatially-Aware Distortions}

\author{
Fadeel Sher Khan\thanks{Work completed during an internship at Samsung Research America.}\\
The University of Texas at Austin\\
Austin, TX, USA\\
{\tt\small fadeelkhan@utexas.edu}
\and
Long N. Le \quad Abhinau K. Venkataramanan\quad Seok-Jun Lee \quad Hamid R. Sheikh\\
Samsung Research America\\
Plano, TX, USA\\
{\tt\small \{longn.le, abhinau.v, seokjun1.lee, hr.sheikh\}@samsung.com}
}

\begin{document}
\maketitle

\begin{abstract}
Traditional image quality assessment (IQA) methods rely on mean opinion scores (MOS), which are resource-intensive to collect and fail to provide interpretable, localized feedback on specific image distortions. We overcome these limitations by shifting from absolute quality prediction to a relational and directional assessment. Our approach utilizes a self-supervised synthetic distortion engine to generate training data, eliminating the need for manual annotation. A distortion prediction network is trained with an anti-symmetric objective to produce spatially-aware, disentangled maps that identify the type, intensity, and direction of distortions relative to a reference image. Subsequently, a scoring network is trained via contrastive learning on ordinally ranked image sets to predict a relational quality score. Our method provides a more granular and interpretable approach to IQA for the targeted optimization of image processing algorithms without requiring any human-labeled quality scores.
\end{abstract}

\section{Introduction}
\label{sec:intro}

Robust image quality assessment (IQA) is a foundational task in computer vision and image processing~\cite{wang2004image}, with critical applications spanning the optimization of computational photography pipelines, the evaluation of generative image synthesis models, automated content moderation, and the benchmarking of image compression algorithms~\cite{Zhai2020}. The primary objective of IQA is to develop computational models that emulate human perception of image quality (IQ)~\cite{wang2004image}. As imaging systems become increasingly complex, incorporating multi-frame fusion, neural image signal processors, and learned post-processing stages, the need for IQA methods that provide not just a score, but actionable and interpretable feedback on specific image characteristics, has become paramount.

Prevailing methodologies for IQA are predominantly reliant on large-scale datasets annotated with human-labeled mean opinion scores (MOS)~\cite{Zhai2020}. While MOS serves as a widely accepted proxy for subjective quality, its utility is constrained by several fundamental factors. First, the process of collecting MOS is resource-intensive. It requires carefully controlled subjective experiments with a statistically significant number of human observers, strict adherence to standardized viewing conditions, and extensive post-processing to screen out unreliable subjects and compute confidence intervals~\cite{bt2002methodology}. Second, MOS is susceptible to substantial inter-rater variability, as individual observers bring different aesthetic preferences, display conditions, and attentional states to the assessment task. Third, and most critically, a single scalar score aggregates diverse image attributes (such as sharpness, noise, color fidelity, compression artifacts, haze, and more) and fails to provide localized, interpretable feedback regarding specific image characteristics or spatial regions.

In this work, we address these limitations by proposing an IQA framework that decouples the objective prediction of image distortions from the subjective scoring of perceptual quality. We reframe the question from absolute quality prediction to a relative and directional assessment, wherein a test image is evaluated relative to a (potentially non-pristine) reference image with respect to its image distortions as well as perceptual quality. We introduce a weakly supervised synthetic distortion engine that generates training triplets (consisting of a reference image, a distorted test image, and a dense ground-truth distortion map) thereby eliminating the need for manual annotation. We subsequently train a distortion prediction network to produce spatially-varying, disentangled maps that identify the type, intensity, and spatial location of specific distortions. By enforcing an anti-symmetric property through its training objective, the network learns to model the directional difference between the image pair, rather than just the magnitude. Finally, a scoring network leverages these explicit distortion maps and is trained via contrastive learning on ordinally ranked image sets to predict a relational quality score defined only by the perceptual ordering of the ranked image sets.
\\ Our primary contributions are:

\begin{itemize}[topsep=0pt,]
    \item a weakly supervised synthetic distortion engine that eliminates the need for human-labeled IQA scores
    \item an IQA network that predicts spatially-varying directional distortion maps
    \item a scoring network that predicts relational IQA using ranked image sets
\end{itemize}

\section{Related Work}
\label{sec:works}

Our work resides in the full reference (FR)-IQA space, \ie, evaluation of a test image against a reference, but draws from self-supervised learning and contrastive learning to address the core limitations of existing methods. We review the most relevant lines of work below.

Traditional FR-IQA metrics are based on simple signal fidelity measures. Peak signal-to-noise ratio (PSNR) quantifies distortion as the mean squared error between a reference and test image, normalized by the dynamic range. While computationally trivial, PSNR operates in the pixel domain and is largely agnostic to the spatial structure and perceptual salience of distortions, leading to well-documented poor correlation with human judgments~\cite{wang2004image}. The structural similarity index (SSIM)~\cite{wang2004image} improved upon this by decomposing the comparison into luminance, contrast, and structural components, yielding substantially better perceptual alignment. Subsequent extensions including multi-scale SSIM (MS-SSIM)~\cite{wang2003multiscale} and information-weighted SSIM (IW-SSIM)~\cite{wang2010information} further improved performance by incorporating multi-resolution analysis and information content weighting. Other traditional approaches include the visual information fidelity (VIF) metric~\cite{sheikh2006image}, which is grounded in natural scene statistics and information-theoretic principles, and the feature similarity index (FSIM)~\cite{zhang2011fsim}, which leverages phase congruency and gradient magnitude features. Despite this, all of these methods produce a single global or locally-averaged scalar that does not extract the contributions of individual distortion types, making them unsuitable for diagnosing \textit{which} specific artifact is degrading quality in \textit{which} spatial region.

Deep learning based IQA methods include the learned perceptual image patch similarity (LPIPS)~\cite{zhang2018unreasonable}, which demonstrated that distances computed in the feature spaces of deep networks trained for image classification (\eg, VGG, AlexNet) correlate remarkably well with human perceptual judgments, often surpassing handcrafted metrics by a significant margin. This insight spawned a family of deep FR-IQA methods that leverage pre-trained or fine-tuned feature extractors to compare reference and test images in a learned representation space. DeepQA~\cite{kim2017deep} proposed learning a quality map from error maps using convolutional neural networks. WaDIQaM~\cite{bosse2017deep} introduced a weighted approach where local quality estimates are aggregated with learned importance weights. More recent methods employ transformer architectures~\cite{golestaneh2022no,qin2023data} to capture long-range dependencies in quality-relevant features.

Most existing methods are trained on large-scale datasets annotated with MOS, such as LIVE~\cite{sheikh2006statistical}, TID2013~\cite{ponomarenko2015image}, KADID-10k~\cite{lin2019kadid}, and PIPAL~\cite{jinjin2020pipal}. The collection of these annotations is costly, and the resulting models are inherently bound to the specific distortion types, intensity ranges, and image content present in the training set, limiting generalization to novel distortion scenarios. These models operate as largely black-box systems wherein they take an image pair as input and produce a single scalar quality prediction without providing actionable interpretability. To mitigate this dependency on absolute MOS annotations, a separate line of research has explored relational IQA, wherein models learn from comparative human judgments rather than absolute scores. The motivation of this work comes from Thurstone's law of comparative judgment~\cite{thurstone1927law}, which posits that humans are more reliable at making pairwise comparisons than at assigning absolute ratings. RankIQA~\cite{liu2017rankiqa} exploits this principle by pre-training a Siamese network on synthetically ranked image pairs before fine-tuning on MOS-labeled datasets, demonstrating that ranking-based pre-training improves MOS prediction. PieAPP~\cite{prashnani2018pieapp} collected a large-scale dataset of pairwise human preference judgments and trained a network to predict the probability that one image is preferred over another, achieving strong perceptual alignment. Dipiq~\cite{ma2017dipiq} generated quality-discriminable image pairs using existing FR-IQA metrics to train a blind IQA model.

These methods represent important steps toward relational assessment, but they still rely on fine-tuning with human-labeled quality scores~\cite{liu2017rankiqa}, or collection of new large-scale human preference annotations~\cite{prashnani2018pieapp}, or leverage existing IQA metrics as pseudo-labels~\cite{ma2017dipiq}. Our method extends the philosophy put forward by this works by generating semantically-aware distortions and learn relational IQA that requires \textbf{\textit{no}} human-labeled IQA scores at any stage.

\section{Proposed Method}
\label{sec:method}

We present our relational IQA framework in three stages. We first formalize the problem setup and the anti-symmetric property that underpins our directional assessment (\cref{sec:proposed_setup}). We then describe the weakly supervised synthetic distortion engine that generates training data without any human annotation (\cref{sec:distortion_engine}). Next, we detail the distortion prediction network architecture and its anti-symmetric training objective (\cref{sec:distortion_network}). Finally, we describe the relational scoring network trained via contrastive learning on ordinally ranked image sets (\cref{ssec:contrastive}). 


\subsection{Problem Formulation}
\label{sec:proposed_setup}

Let the space of all possible images be denoted by $\mathcal{I} \in \mathbb{R}^{C \times H \times W}$, where $C$, $H$, and $W$ represent the number of color channels, height, and width, respectively. Let a reference image $I_{\text{ref}} \in \mathcal{I}$ serve as the conditional baseline for evaluating a test image $I_{\text{test}} \in \mathcal{I}$. The purpose of the reference image is to provide a specific, contextual frame of reference from which image distortions in $I_{\text{test}}$ are measured and defined. We do not impose any spatial constraints requiring $I_{\text{test}}$ and $I_{\text{ref}}$ to be pixel-aligned, and $I_{\text{ref}}$ does not have to be pristine. This is important as the reference may itself be the output of some processing network and thus non-pristine in some image attributes (\eg, worse image quality in some regions as compared to the test image). Let the space of spatially-varying distortion maps be $\mathcal{D} = [0, 1]^{N \times H \times W}$, where $N$ is the number of distinct distortion types under consideration. Each channel of a distortion map corresponds to a specific distortion category (\eg, blur, noise, haze), and the value at each spatial location encodes the intensity of that distortion.

Our objective is to learn a mapping $\mathcal{F}: \mathcal{I} \times \mathcal{I} \to \mathcal{D}$ such that for any pair $(I_{\text{test}}, I_{\text{ref}})$, the output $I_{\text{distortion}} = \mathcal{F}(I_{\text{test}}, I_{\text{ref}})$ represents the distortions present in $I_{\text{test}}$ relative to $I_{\text{ref}}$. We model $\mathcal{F}$ as an approximation of a conditional probability distribution. Let $D$ be a random variable representing an image distortion of some intensity at a given spatial location. The output of our function at each spatial coordinate $(h, w)$ is a vector in $\mathbb{R}^N$ that approximates the posterior probability of each distortion type, conditioned on the image pair:
\begin{equation}
\mathcal{F}(I_{\text{test}}, I_{\text{ref}})_{[:, h, w]} \approx P(D | I_{\text{test}}[h, w], I_{\text{ref}}[h, w])
\end{equation}

We design $\mathcal{F}$ to be anti-symmetric with respect to the ($I_{\text{test}}, I_{\text{ref}}$) pair, i.e., it models distortions in $I_{\text{test}}$ \textit{relative} to $I_{\text{ref}}$ and thus $ \mathcal{F}(I_{\text{test}}, I_{\text{ref}}) = 1- \mathcal{F}(I_{\text{ref}}, I_{\text{test}})$. This is in contrast to symmetric methods (for example, $L_p$ norms) that model a simpler distance estimation or joint probability and thus only quantify the \textit{magnitude} of the difference, not its \textit{direction}. We constrain $\mathcal{F}$ as such to better align with perceptual IQA. Humans perceive the quality of one image with respect to another better than standalone \cite{prashnani2018pieapp}, i.e., conditioning, and are highly sensitive to local variations in IQ \cite{Gerhard2013}. By this, $I_{\text{test}}$ can locally be better than $I_{\text{ref}}$ and vice versa, and our method should thus able to capture this. Symmetric methods are directionally blind so they do not satisfy this property.

\subsection{Semantic-Aware Synthetic Distortion Engine}
\label{sec:distortion_engine}

A key challenge in training a dense distortion prediction network is obtaining ground-truth annotations of distortions. We address this through a weakly supervised synthetic distortion engine that automatically generates training triplets $(I_{\text{ref}}, I_{\text{test}}, Y)$ with dense ground-truth distortion maps without the need for any IQA scalar scores.

Given a reference image $I_{\text{ref}}$, a pre-trained semantic segmentation network produces a set of $K$ mutually exclusive binary masks $\{m_k\}_{k=1}^K$, where $m_k \in \{0, 1\}^{H \times W}$ and $\sum_{k=1}^K m_k = \mathbf{1}_{H \times W}$. Each mask corresponds to a distinct semantic region (\eg, sky, vegetation, building, person). The use of semantic masks, rather than random spatial regions, is a deliberate design choice motivated by the observation that real-world image processing artifacts often manifest differently across semantic categories. For example, a multi-frame noise reduction algorithm may effectively denoise flat sky regions while introducing blur in high-frequency texture regions such as foliage or fabric. By connecting distortions to semantic boundaries, our synthetic data more closely mirrors the spatial structure of artifacts encountered in real-life settings. To increase diversity and prevent the network from overfitting to semantic boundaries alone, we also include random binary masks (\ie, generated via thresholded Perlin noise \cite{perlin1985} or random rectangular regions) in the mask pool with some probability during training.

We define a bank of $N$ distortion transformation operators $\mathcal{X} = \{x_1, \dots, x_N\}$, each modulated by an intensity scalar $\alpha$ (see \cref{sec:experiments} for specific distortions used). For each semantic region $m_k$, we independently sample a distortion operator $x_{j_k} \in \mathcal{X}$ and an intensity scalar $\alpha_k \in [0, 1]$. The distorted test image $I_{\text{test}}$ is synthesized by compositing the individually distorted regions:
\begin{equation}
\label{eq:itest}
I_{\text{test}} = \sum_{k=1}^K m_k \odot x_{j_k}(I_{\text{ref}}, \alpha_k)
\end{equation}
where $\odot$ denotes the element-wise (Hadamard) product. When $\alpha_k = 0$, the distortion operator acts as the identity, leaving the region unchanged. This allows some semantic regions to remain undistorted, mimicking the common scenario where artifacts are localized to specific image areas.

Concurrently with the distorted image, we generate a dense ground-truth distortion map $Y \in \mathbb{R}^{N \times H \times W}$. Each channel $Y^{(j)}$ of this map corresponds to the distortion operator $x_j$ and encodes the intensity of that distortion at each spatial location:
\begin{equation}
Y^{(j)} = \sum_{k=1, j_k=j}^K \alpha_k \cdot m_k
\end{equation}
By construction, the distortion map $Y$ is sparse. At each $(h, w)$, at most one channel has a non-zero value (corresponding to the single distortion applied to that region). This sparsity reflects the assumption that within a given semantic region, a single dominant distortion type is present, though different regions may exhibit different distortion types. The resulting training triplet $(I_{\text{ref}}, I_{\text{test}}, Y)$ is generated entirely without manual annotation, where $I_{\text{test}}$ is the locally distorted version of $I_{\text{ref}}$ and $Y$ provides a dense, disentangled annotation of the spatially-varying distortions.

To ensure the distortion prediction network generalizes beyond the specific distortion operators in $\mathcal{X}$, we employ several strategies to maximize the diversity of the synthetic training data. Intensity scalars $\alpha_k$ are sampled from a configurable distribution (uniform or beta) to cover a wide range of severity levels. The number of distorted regions $K$ is varied across samples, and the spatial scale of masks ranges from large semantic regions covering significant portions of the image to small localized patches, ensuring the network learns to distinguish true distortions from benign photometric variations.

\subsection{Distortion Prediction Network}
\label{sec:distortion_network}

We train $\mathcal{F}_\theta(I_A, I_B)$ parameterized by $\theta$ to predict the dense distortion map $\hat{Y} \in \mathbb{R}^{N \times H \times W}$. We adapt the Mask2Former architecture proposed in~\cite{cheng2022masked}, originally designed for universal image segmentation via semantic mask prediction, to output continuous-valued spatial regression maps rather than discrete class predictions. The architecture consists of a shared backbone feature extractor (\eg, a Swin Transformer~\cite{liu2021swinv2}) which processes the input images to produce multi-scale feature maps. We adapt this for FR-IQA by concatenating the feature representations of $I_{\text{test}}$ and $I_{\text{ref}}$ along the channel dimension, enabling the network to reason about the relationship between the two images in the feature space rather than the pixel space. This feature-level fusion strategy allows the network to leverage high-level semantic information from both images when predicting distortions. A pixel decoder progressively upsamples and refines the multi-scale features into a high-resolution feature map. A transformer decoder employs a set of $N$ learnable queries $\{q_n\}_{n=1}^N$, one per distortion type. Each query $q_n$ is specialized to detect a specific distortion type $d_n$ via masked cross-attention over the fused image features. The masked cross-attention mechanism restricts each query to attend only to spatial regions where the corresponding distortion is predicted to be present, promoting distortion-specific specialization and spatial disentanglement. The final output $\hat{Y} = \mathcal{F}_\theta(I_A, I_B)$ is formed by projecting each query's output back into a spatial map via a dot product with the per-pixel embeddings from the pixel decoder, followed by a sigmoid activation to constrain outputs to $[0, 1]$.

To train $\mathcal{F}_\theta$, we minimize a \textit{weighted} mean square error between the predicted map $\hat{Y}$ and the ground truth distortion map $Y$. Because the ground-truth maps are inherently sparse (most pixels have zero distortion in most channels), a naive MSE loss would regress towards zero. To counteract this class imbalance, we apply a spatially-varying weight map $W$ that assigns a substantially higher penalty to errors on pixels where the ground truth intensity is non-negligible (\ie, exceeds a threshold $\beta$):
\begin{equation}
W_{ij} =
\begin{cases}
  w_{\text{high}} & \text{if } |Y_{ij}| > \beta \\
  1 & \text{otherwise}
\end{cases}
\end{equation}
where $w_{\text{high}} \gg 1$. This ensures the network accurately predicts the intensity and spatial extent of active distortion regions, which are the meaningful portions of the map.

To enforce the anti-symmetric property $ \mathcal{F}(I_{\text{test}}, I_{\text{ref}}) = 1- \mathcal{F}(I_{\text{ref}}, I_{\text{test}})$, we employ a stochastic anti-symmetric training objective. During training, for each triplet $(I_{\text{test}}, I_{\text{ref}}, Y)$ in a minibatch, we introduce a probability $p_{\text{swap}}$ of swapping the input pair to $(I_{\text{ref}}, I_{\text{test}})$ and concurrently replacing the ground-truth map with $1-Y$. This stochastic swapping, rather than deterministic augmentation of every sample, exposes the network to both orderings with different frequencies, encouraging it to learn the directional difference as an intrinsic property of its representation rather than memorizing the input ordering. The training objective over a minibatch is:
\begin{equation}
\label{weighted_mse}
\begin{split}
\mathcal{L}_{\text{mse}} = & \sum_{(I_A, I_B, Y)} \| \mathcal{F}_\theta(I_A, I_B) - Y \|^2_W + \\
& \sum_{(I_A, I_B, Y)} \| \mathcal{F}_\theta(I_B, I_A) - (1-Y) \|^2_W
\end{split}
\end{equation}
The first term trains the network to predict the forward distortion map, while the second term explicitly trains the reverse mapping, enforcing that $\mathcal{F}_\theta(I_A, I_B) \approx 1- \mathcal{F}_\theta(I_B, I_A)$. Together, these terms ensure that the learned representation is inherently directional, and the network does not merely detect that a distortion is present, but encodes which image in the pair exhibits the distortion relative to the other.

\subsection{Relational IQA via Contrastive Learning}
\label{ssec:contrastive}

The distortion prediction network trained in \cref{sec:distortion_network} produces interpretable, spatially-aware distortion maps but does not directly yield a quality score. To bridge the gap between distortion characterization and subjective quality assessment, we train a second-stage scoring network that learns to map distortion maps and images to a relational quality score. This scoring network does not require absolute MOS labels, but learns only from the ordinal ranking of image sets (\ie, weak supervision).

We consider a collection of image sets, denoted by $\mathcal{I} = \{\mathcal{I}_t\}_{t=-T}^{0}$, where each set $\mathcal{I}_t = \{I_t^{(j)}\}_{j=1}^{n}$ contains $n$ images. We assume that all images within a given set $\mathcal{I}_t$ share the same perceptual quality tier, and that the image quality (IQ) is monotonically ordered across these sets:
$$
IQ(\mathcal{I}_{-T}) < IQ(\mathcal{I}_{-T+1}) < \dots < IQ(\mathcal{I}_{0})
$$
This ordinal structure can be obtained far more cheaply than MOS. It requires only a coarse ranking of image sets by overall quality, without assigning numerical scores to individual images. In our setting, this ranking arises naturally from the experimental design (processing images with networks of decreasing capacity), supplemented by lightweight visual inspection.

The scoring network takes as input an image $I_t^{(j)}$ concatenated with its associated predicted distortion map $\hat{Y}_t^{(j)} = \mathcal{F}_\theta(I_t^{(j)}, I_{\text{ref}})$ along the channel dimension, forming an (RGB+$N$)-channel input tensor. This input is processed by a pre-trained Swin Transformer~\cite{liu2021swinv2} backbone whose first convolutional layer is modified to accept the expanded channel count. The backbone produces a feature embedding $F_t^{(j)} \in \mathbb{R}^{D}$, which is then passed through a lightweight MLP head to produce a scalar quality score $S_t^{(j)}$. By conditioning the scoring network on the explicit distortion maps, we provide it with structured, interpretable information about the distortion characteristics, rather than requiring it to implicitly re-learn distortion detection from raw pixels.

We train the scoring network using a contrastive training objective. Specifically, we utilize a hinge loss to enforce the ordinal relationship between the scores of consecutively ranked image sets:
\begin{equation}
\mathcal{L}_{\text{hinge}} = \max(0, \delta - (S_{t+1}^{(j)} - S_t^{(j)})),
\label{eq:hinge}
\end{equation}
where $\delta$ is a predefined margin. This loss ensures that the score $S_{t+1}^{(j)}$ of a higher-quality image exceeds the score $S_t^{(j)}$ of a lower-quality version of image $j$ by at least $\delta$. By operating on consecutive pairs in the ranking, the hinge loss collectively enforces the full ordinal structure.

In addition to the ordinal score loss, we structure the feature embedding space using a contrastive objective based on the InfoNCE loss~\cite{oord2019representationlearningcontrastivepredictive}. The goal is to ensure that feature embeddings of images with the same perceived quality tier are grouped together in the embedding space, while embeddings of images from different quality tiers are pushed apart. For an anchor image $I_t^{(i)}$, we define positive samples as all other images from the same quality tier $\{I_t^{(j)}\}_{j \neq i}$ and negative samples as all images from all other quality tiers $\{I_{t'}^{(k)}\}_{t' \neq t, \forall k}$. The contrastive loss for the anchor embedding $F_t^{(i)}$ is:
\begin{equation}
\label{eq:infonce}
\mathcal{L}_{IN} = -\mathbb{E}_{F_t^{(i)}} \left[ \log \frac{
  \sum_{j \neq i} \exp(\text{sim}(F_t^{(i)}, F_t^{(j)}) / \tau)
}{
  \sum_{(t',k) \neq (t,i)} \exp(\text{sim}(F_t^{(i)}, F_{t'}^{(k)}) / \tau)
} \right]
\end{equation}
where $\text{sim}(\cdot, \cdot)$ denotes cosine similarity and $\tau$ is a temperature hyperparameter that controls the sharpness of the distribution. The InfoNCE loss complements the hinge loss by providing a richer training signal. While the hinge loss only constrains the scalar score ordering, the InfoNCE loss structures the entire feature embedding space, encouraging the network to learn representations that capture meaningful quality-related features beyond what is needed for ordinal scoring alone. This structured embedding space can be leveraged for downstream tasks such as quality-based retrieval or clustering.

Our net loss function for the scoring network combines the ordinal and contrastive objectives:
\begin{equation}
\label{eq:total_loss}
    \mathcal{L} = \lambda_{\text{rank}}\mathcal{L}_{\text{hinge}} + \lambda_{\text{con}}\mathcal{L}_{IN}
\end{equation}
where $\lambda_{\text{rank}}$ and $\lambda_{\text{con}}$ are scalar weights that balance the two loss components. Through this formulation, we do not need labeled subjective MOS for any image (or any image set) but rather only a relativistic, unscored ranking that lets us learn a perceptual direction to our IQA scoring. The distortion prediction network's parameters are frozen during this stage, and only the scoring network is trained.
\begin{figure*}[t]
    \centering
    \includegraphics[width=\textwidth]{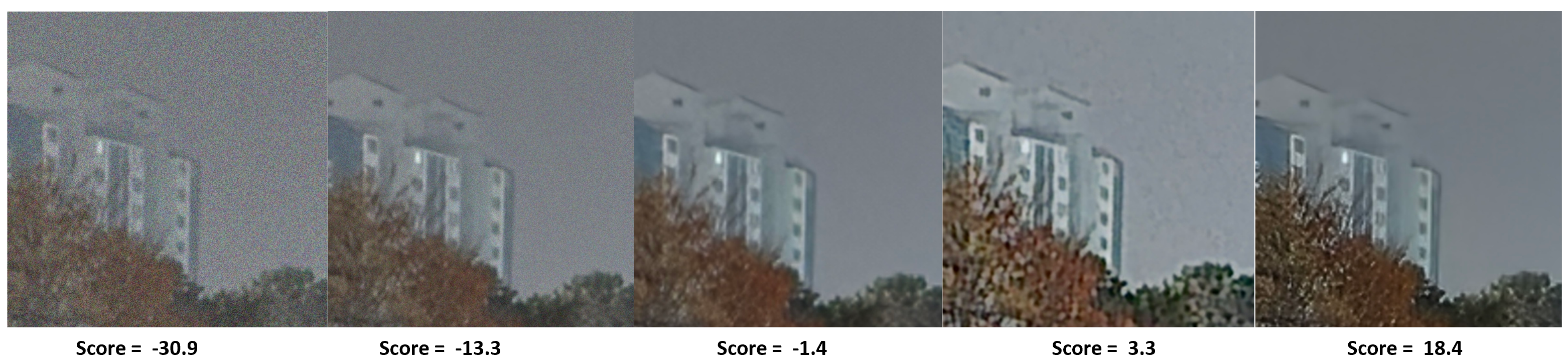}
    \caption{Relational IQA scoring correlates with lower levels of global and localized distortions. In this example, images are produced by models of increasing network capacity and differ in IQ. The image with highest score has more textured patterns in semantic regions (\ie, higher resolution), lower uniform and non-uniform noise, and less haze artifacts.}
    \label{fig:visual_results_scoring}
\end{figure*}

\begin{figure}[t]
  \centering
  \includegraphics[width=\linewidth]{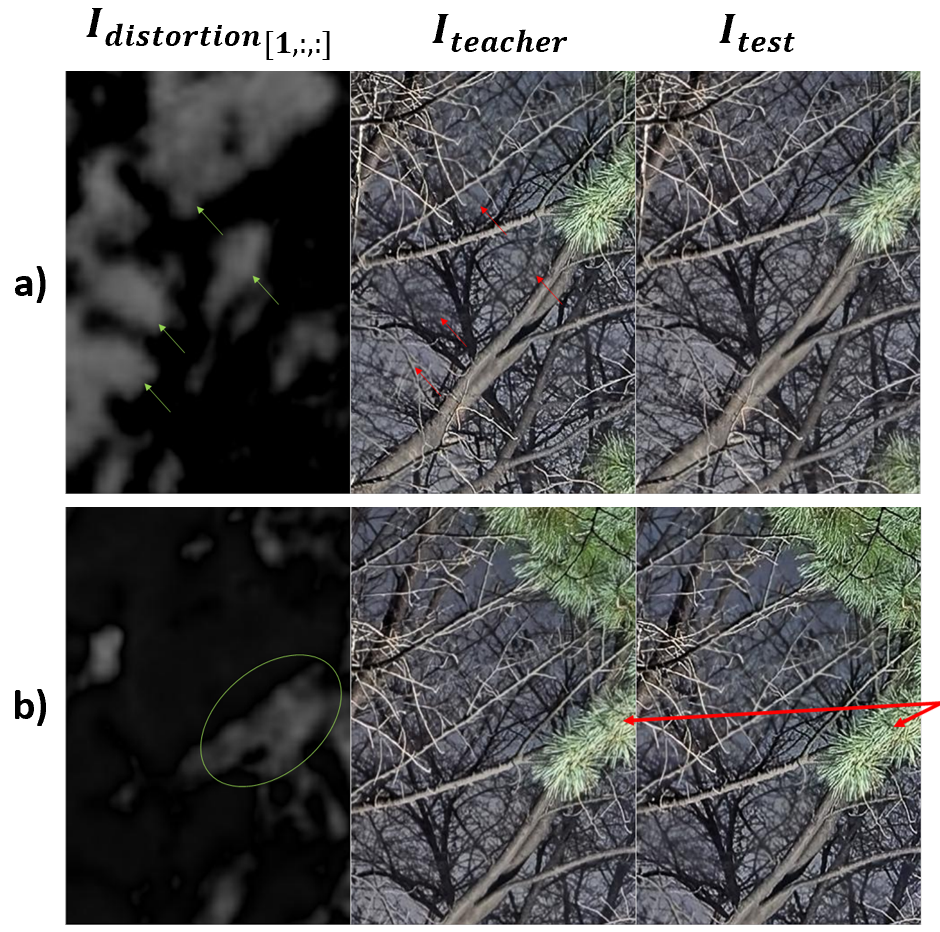}
    \caption{\textbf{(a)} Distortion map $\hat{Y} \in \mathbb{R}^{1\times H \times W}$ of the channel representing local blur in $I_{test}$ relative to $I_{ref}$. \textbf{(b)} Images from \textbf{(a)} are flipped and thus $\hat{Y}$ is now able to show blurry regions in new $I_{test}$ relative to the new $I_{ref}$ owing to anti-symmetric training (\cref{sec:method}). Local blur in $I_{test}$ indicated by \textcolor{red}{red} arrows and corresponding regions on $\hat{Y}$ indicated by \textcolor{green}{green} arrows.}
\label{fig:distortion_map}
\end{figure}

\section{Experiments}
\label{sec:experiments}

\subsection{Setup}
\label{sec:dataset}

We utilize a proprietary dataset of 270 handheld multi-frame mobile phone captures at 12 megapixel resolution. Each capture consists of multiple RAW Bayer frames acquired in rapid succession under natural lighting conditions, spanning a diverse range of indoor and outdoor scenes, lighting conditions (from bright daylight to low-light), and semantic content. Images are processed from the Bayer color space to RGB using an existing multi-frame image signal processing (ISP) network~\cite{Khan_2025_CVPR} that performs demosaicing, denoising, multi-frame alignment and fusion, tone mapping, and color correction. From each processed 12-megapixel image, we extract 200 non-overlapping patches of size $512 \times 512$ pixels in the RGB space, yielding a total of 54,000 source patches.

For this study, our synthetic distortion engine employs a bank of $N=6$ distortion transformation operators, each designed to emulate common artifacts encountered in computational photography. \textit{Gaussian blur} simulates resolution loss and defocus by convolving the image with a Gaussian kernel whose standard deviation is modulated by $\alpha$. \textit{Structured noise (Perlin noise)}~\cite{perlin1985} generates spatially correlated noise patterns that resemble sensor-induced fixed-pattern noise or banding artifacts, with amplitude controlled by $\alpha$. \textit{Checkerboard patterns} simulates demosaicing artifacts and aliasing by superimposing periodic checkerboard patterns at varying frequencies and amplitudes. \textit{Bad pixels} introduces isolated or clustered pixel-level defects (\ie, dead pixels, hot pixels) at random locations within the masked region, with the density of defective pixels controlled by $\alpha$. \textit{Haze} simulates atmospheric haze or lens flare by blending the image region toward a uniform gray or white value, with the blending factor determined by $\alpha$. \textit{Over-saturation} increases the color saturation of the image region beyond natural levels, controlled by $\alpha$, simulating aggressive color processing or white balance errors.

Each distortion operator $x_j$ accepts a reference image patch and an intensity scalar $\alpha_k \in [0, 1]$ as inputs, where $\alpha_k = 0$ corresponds to no distortion (identity) and $\alpha_k = 1$ corresponds to maximum distortion severity. These operators are applied to localized semantic and random regions sampled from $I_{\text{ref}}$ as described in \cref{sec:distortion_engine}, generating the training triplets $(I_{\text{ref}}, I_{\text{test}}, Y)$.

For semantic mask generation, we employ a pre-trained DeepLabV3+~\cite{chen2018encoder} model with a ResNet-101 backbone trained on the ADE20K dataset~\cite{zhou2019semantic}. This produces per-pixel semantic labels which are grouped into $K$ regions. We supplement semantic masks with random binary masks generated via thresholded Perlin noise with probability $0.3$ to increase spatial diversity.

To create the ordinally ranked image sets required for contrastive learning (\cref{ssec:contrastive}), we process the same RAW Bayer captures through a previously defined multi-frame ISP network~\cite{Khan_2025_CVPR} at decreasing levels of network capacity by systematically reducing the number of parameters ($n_{t=0} > n_{t=-1} > \dots > n_{t=-T}$). Reducing network capacity degrades the ISP's ability to perform high-fidelity denoising, sharpening, and tone mapping, resulting in images with progressively lower perceptual quality while preserving the same scene content and composition. This yields a natural ordinal ranking of image quality tied to network capacity.

To ensure the assumed ordinal quality constraint $IQ(\mathcal{I}_{-T}) < IQ(\mathcal{I}_{-T+1}) < \dots < IQ(\mathcal{I}_{0})$ holds reliably, image sets are visually inspected and pruned by a panel of $n_{\text{inspectors}}=4$ trained image quality analysts. Images where the ordinal assumption is violated (\eg, due to scene-specific artifacts where a smaller network coincidentally produces a more pleasing result) are removed from the dataset. This quality control step ensures a clean ordinal signal for the contrastive learning stage.

Additionally, the image set with the lowest IQ (\ie, $\mathcal{I}_{-T}$) is taken and a distortion schedule is applied with increasing synthetic distortion intensity $\alpha$ (\cref{eq:itest}) to create further degraded image sets. This extends the quality range beyond what can be achieved through network capacity reduction alone, providing additional training signal at the low-quality end of the spectrum. For this study, we analyze 6 image sets created by multi-frame Bayer-to-RGB networks with decreasing network capacity (2 of these sets are reserved for validation) and an additional 3 image sets with increasing synthetic distortions, yielding $T=9$ quality tiers in total. The image set with the highest IQ (\ie, $\mathcal{I}_{t=0}$) serves as the reference image set $I_{\text{ref}}$ for all $\mathcal{I}_{t\neq 0}$ and is thus excluded from the relational scoring dataset to avoid biasing the learned score toward the reference.

\subsection{Implementation}
\label{sec:implementation}

As described, the distortion prediction network adapts the Mask2Former~\cite{cheng2022masked} architecture. The backbone is a Swin Transformer-Large~\cite{liu2021swinv2} pre-trained on ImageNet-22k. We utilize the transformer decoder with embedding dimension of $1024$, $6$ decoder layers (each with $16$ attention heads), and $N=6$ learnable distortion queries corresponding to the $N$ distortion types. The pixel decoder employs a multi-scale deformable attention mechanism following~\cite{cheng2022masked}. The network is trained with the anti-symmetric objective (\cref{weighted_mse}) using a swap probability $p_{\text{swap}} = 0.25$, weight map threshold $\beta = 0.05$, and focal weight $w_{\text{high}} = 10$.

The scoring network uses a pre-trained Swin Transformer-Base~\cite{liu2021swinv2} backbone, with the input patch embedding layer modified to accept (RGB+$N$)-channel inputs (3 RGB channels + 6 distortion map channels = 9 channels). The additional input channels are initialized with zero weights to preserve the pre-trained backbone's behavior at initialization. The embedding dimension is $D=768$, and the MLP scoring head consists of two fully-connected layers with GELU activation and a final linear projection to a scalar.

All training is conducted on $8\times$ NVIDIA V100 GPUs using the AdamW optimizer~\cite{loshchilov2018decoupled} with a batch size of $8$ per GPU and a base learning rate of $5\times10^{-5}$ with cosine annealing. The distortion prediction network is trained for $600$ epochs, and the relational scoring network is subsequently trained for $300$ epochs with the distortion network's parameters frozen. We set the scoring hyperparameters to $\lambda_{\text{rank}}=1.0$, $\lambda_{\text{con}}=0.5$, hinge margin $\delta=1.0$, and InfoNCE temperature $\tau=0.07$. Standard data augmentation is applied during training, including random horizontal and vertical flips, random rotations ($0^\circ$, $90^\circ$, $180^\circ$, $270^\circ$), and random crops. For the distortion prediction network, augmentations are applied identically to both $I_{\text{ref}}$ and $I_{\text{test}}$ (and the ground-truth map $Y$ is augmented consistently) to preserve spatial correspondence.

\subsection{Results}
\label{sec:results}

To our knowledge, there do not exist directly comparable methods or publicly accessible datasets of the form $\mathcal{I} = \{\{I_t^{(j)}\}_{j=1}^{n} \in \mathcal{I}_t\}_{t=-T}^{0}$ suitable for evaluating relational IQA scoring. Existing benchmarks are designed around absolute MOS prediction and evaluate metrics such as Spearman rank-order correlation coefficient (SRCC) and Pearson linear correlation coefficient (PLCC) against ground-truth MOS values. Since our framework does not predict absolute MOS, direct comparison with MOS-based methods on standard benchmarks is not meaningful. Instead, we evaluate our method on its ability to maintain correct ordinal rankings and produce perceptually consistent scores alongside qualitative evaluation.

\Cref{fig:visual_results_scoring} presents relational scores across image sets of increasing quality, demonstrating that our scores monotonically increase with decreasing distortion levels. The images produced by the highest-capacity network exhibit more textured patterns in semantic regions (\ie, higher effective resolution), lower uniform and non-uniform noise, and fewer haze artifacts, all of which are correctly reflected in the relational scores. Notably, the scoring network generalizes to the validation image sets (unseen during training), correctly ranking them within the quality hierarchy established by the training sets.

 \Cref{fig:distortion_map} illustrates the anti-symmetric distortion map prediction for the blur channel. In panel (a), the distortion map $\hat{Y}$ correctly identifies regions in $I_{\text{test}}$ that are blurrier than the corresponding regions in $I_{\text{ref}}$, as indicated by high activation values. When the input pair is swapped in panel (b), the distortion map inverts: the previously highlighted regions now show low activation (indicating that the new $I_{\text{test}}$, which was the former $I_{\text{ref}}$, is sharper in those regions), while new regions are highlighted where the new $I_{\text{test}}$ exhibits blur. This behavior confirms that the anti-symmetric training objective successfully teaches the network to model directional quality differences rather than merely detecting the presence of blur.

The spatial precision of the predicted distortion maps is notable. The maps exhibit clean boundaries that align well with both semantic edges (\eg, transitions between foreground and background) and distortion boundaries (\eg, the spatial extent of a locally applied blur). This spatial precision is a direct consequence of the semantic-aware synthetic distortion engine, which generates training data with distortions tied to meaningful spatial structures.

\section{Discussion}
\label{sec:discussion}

Our framework offers several practical advantages over conventional MOS-based IQA. First, the weakly supervised synthetic distortion engine eliminates the most expensive component of IQA system development: large-scale human annotation. The ordinal ranking required for the scoring network is vastly cheaper to obtain than absolute MOS, as it requires only coarse visual inspection rather than controlled subjective experiments. Second, the spatially-aware, distortion-disentangled maps provide actionable diagnostic information that a single scalar score cannot. An imaging pipeline engineer can examine the distortion maps to identify that, for example, the multi-frame fusion stage is introducing haze in sky regions or that the sharpening stage is exacerbating noise in dark regions. This level of specificity enables targeted, efficient optimization. Third, the relational formulation naturally supports non-pristine references, enabling comparison between the outputs of competing algorithms or different configurations of the same system.

Our current framework also has some limitations that suggest directions for future research. The distortion bank $\mathcal{X}$ is fixed and finite. While the six distortion types we employ cover many common artifacts, extending to a more comprehensive set would broaden applicability. Learning the distortion operators themselves in a data-driven manner is an intriguing direction. The relational scoring currently requires that all images in a given quality tier share the same scene content, and extending to cross-scene relational comparisons would further increase the method's generality. Additionally, while our method is demonstrated on computational photography data, evaluating on standard IQA benchmarks by learning a mapping from our relational scores to MOS would enable direct comparison with existing methods.

Beyond quality assessment, our framework opens several application pathways. The spatially-aware distortion maps (\cref{fig:distortion_map}) can serve as a dense reward signal for reinforcement learning-based optimization of image processing networks, or as a spatially-varying perceptual loss function that penalizes specific distortion types in specific regions during training of generative or restoration models. The ability to score ordinally ranked image sets (\cref{fig:visual_results_scoring}) can facilitate automated model checkpoint selection, \ie during neural network training, intermediate checkpoints can be ranked by our relational scoring to identify the point of best perceptual quality, defined only by the perceptual direction encoded in the ranked training sets $\{\mathcal{I}_t\}_{t=-T}^{0}$, without requiring human evaluation of each checkpoint. Furthermore, the structured embedding space learned by the InfoNCE loss could support quality-based image retrieval and clustering applications.
\section{Conclusion}
\label{sec:conclusion}

We present a relational IQA framework that bypasses the need for expensive human opinion scores by using a weakly supervised synthetic distortion engine. Our method predicts interpretable, spatially-aware directional distortion maps. Further training with contrastive learning on ordinally ranked image sets produces a relational image quality score. Future work will explore its applicability for model training, checkpointing, and evaluation.

{
    \small
    \bibliographystyle{ieeenat_fullname}
    \bibliography{main}
}

\end{document}